\documentclass[conference]{IEEEtran}

\IEEEoverridecommandlockouts

\usepackage{mathrsfs}
\usepackage{amsmath}
\usepackage{amsfonts}
\usepackage{amsthm}
\usepackage{amssymb}
\usepackage{graphicx}
\usepackage{subfigure}
\usepackage{indentfirst}
\usepackage{array}
\usepackage{epstopdf}
\usepackage{cite}
\usepackage{enumerate}
\usepackage{bm}
\usepackage{dsfont}
\usepackage[linesnumbered,ruled,vlined]{algorithm2e}
\usepackage{multirow}
\usepackage{verbatim}
\usepackage{stfloats}
\usepackage{color}

\SetKwComment{Comment}{//}{}
\SetKwBlock{Begin}{Function}{end}

\begin{document}
	
\title{Perceptual Learned Source-Channel Coding for High-Fidelity Image Semantic Transmission}	
	
\author
{
\IEEEauthorblockN{Jun~Wang\IEEEauthorrefmark{1},  Sixian~Wang\IEEEauthorrefmark{1},
Jincheng~Dai\IEEEauthorrefmark{1},
Zhongwei~Si\IEEEauthorrefmark{1}, Dekun~Zhou\IEEEauthorrefmark{1}, and Kai~Niu\IEEEauthorrefmark{1}\IEEEauthorrefmark{2}}
\IEEEauthorblockA{\IEEEauthorrefmark{1}Key Laboratory of Universal Wireless Communications, Ministry of Education}
\IEEEauthorblockA{\IEEEauthorrefmark{1}Beijing University of Posts and Telecommunications, Beijing, China}
\IEEEauthorblockA{\IEEEauthorrefmark{2}Peng Cheng Laboratory, Shenzhen, China}

\thanks{This work was supported in part by the National Natural Science Foundation of China under Grant 92067202, Grant 62001049, and Grant 61971062, in part by the Beijing Natural Science Foundation under Grant 4222012. (\emph{Corresponding author: Jincheng Dai}, email: daijincheng@bupt.edu.cn)}

\vspace{0em}

}
\maketitle

\begin{abstract}
As one novel approach to realize end-to-end wireless image semantic transmission, deep learning-based joint source-channel coding (deep JSCC) method is emerging in both deep learning and communication communities. However, current deep JSCC image transmission systems are typically optimized for traditional distortion metrics such as peak signal-to-noise ratio (PSNR) or multi-scale structural similarity (MS-SSIM). But for low transmission rates, due to the imperfect wireless channel, these distortion metrics lose significance as they favor pixel-wise preservation. To account for human visual perception in semantic communications, it is of great importance to develop new deep JSCC systems optimized beyond traditional PSNR and MS-SSIM metrics. In this paper, we introduce adversarial losses to optimize deep JSCC, which tends to preserve global semantic information and local texture. Our new deep JSCC architecture combines encoder, wireless channel, decoder/generator, and discriminator, which are jointly learned under both perceptual and adversarial losses. Our method yields human visually much more pleasing results than state-of-the-art engineered image coded transmission systems and traditional deep JSCC systems. A user study confirms that achieving the perceptually similar end-to-end image transmission quality, the proposed method can save about 50\% wireless channel bandwidth cost.
\end{abstract}

\section{Introduction}

\begin{figure}[t]
	\setlength{\abovecaptionskip}{0.cm}
	\setlength{\belowcaptionskip}{-0.cm}
	\centering{\includegraphics[scale=0.56]{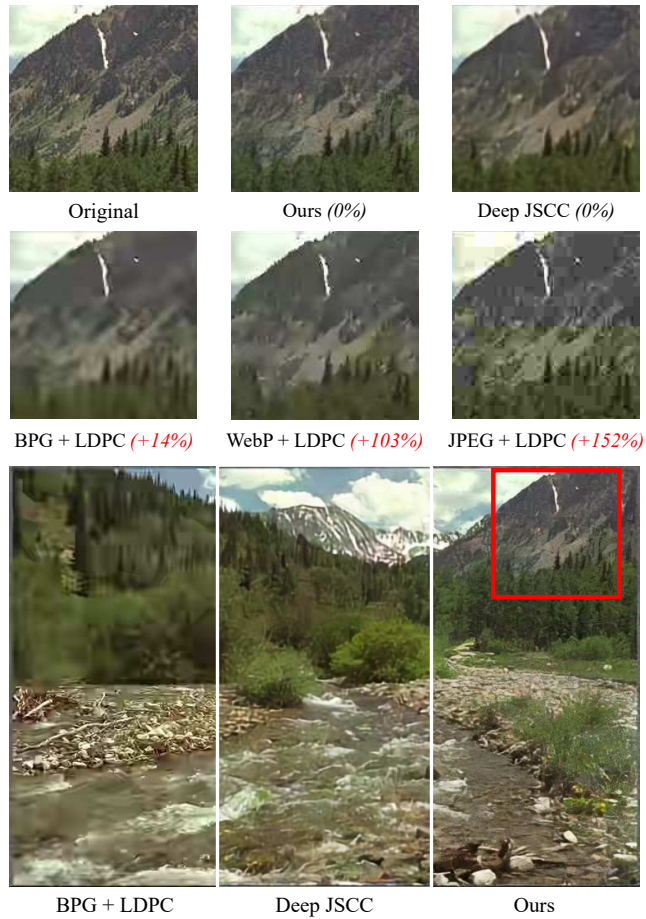}}
	\caption{Visual comparison of our new perceptually learned deep JSCC to that obtained by other image coded transmission systems. All schemes are transmitted over additive white Gaussian noise (AWGN) channel at signal-to-noise ratio (SNR) 10dB. Red number indicate the percentage of wireless channel bandwidth cost increase compared to our proposed scheme (channel bandwidth ratio (coding rate) is 1/48). Note that even when using equivalent or much more channel bandwidth, all other coded transmission schemes are outperformed by our method visually.}\label{Fig_introduction_demo}
\vspace{-1em}
\end{figure}

End-to-end wireless image transmission systems based on deep joint source-channel coding (deep JSCC) are becoming an active area of research \cite{DJSCC,DJSCCF}. Compared to traditional image coded transmission systems, such as BPG image compression (the state-of-the-art (SOTA) engineered image compression codec \cite{BPG}) combined with 5G low-density parity-check (LDPC) channel coded modulation \cite{LDPC_5G}, deep JSCC shows competitive performance due to its source image semantics-aware and channel-aware abilities. Despite this, current deep JSCC image transmission systems are typically optimized under traditional distortion metrics such as peak signal-to-noise ratio (PSNR) or multi-scale structural similarity (MS-SSIM) \cite{msssim}. These simple and shallow metrics fail to account for human perception in terms of many image nuances. Especially for low transmission rates, due to the imperfect wireless channel, traditional distortion metrics, such as mean-squared error (MSE) and MS-SSIM, lose significance as they favor pixel-wise preservation \cite{agustsson2019}. This is apparently not aligned with human perception in semantic communications, which tend to focus on global semantic information and local texture. To tackle this, it is of great importance to account for human perception during the optimization of deep JSCC systems.

Recent deep learning techniques in computer vision suggest that perceptual metric and adversarial training strategy play important roles in image compression tasks that can well improve the visual quality of reconstructed images. As for the image quality assessment, Zhang et al. \cite{lpips} has proposed the learned perceptual image patch similarity (LPIPS) model, measuring the Euclidean distance between deep representations. It has been shown that feature maps of different neural network architectures have ``reasonable'' effectiveness in accounting for human perception of image quality. LPIPS has been widely identified as a generalization of the ``perceptual loss''. To improve human visual quality, adversarial losses were employed in image compression tasks \cite{agustsson2019,HIFIC,rippel2017real,santurkar2018generative}. Motivated by these emerging techniques to augment reconstructed image visual quality, it is the very time to design new deep JSCC system to coincide with human perception, which is consistent with the goal of semantic communications.

In this paper, we propose a new learned deep JSCC method for end-to-end wireless transmission of full-resolution images. Our method integrates both perceptual loss and adversarial loss to capture global semantic information and local texture thus yielding deep JSCC systems that produce human virtually appealing images. In contrast to SOTA engineered image coded transmission systems and recently proposed deep JSCC systems, our new deep JSCC system combines encoder, wireless channel, decoder/generator, and discriminator together, which can be jointly trained under both perceptual and adversarial losses. Our proposed end-to-end wireless image transmission system tends to preserve the full image, also falling back to the synthesized content instead of blocky/blurry regions caused by the limited channel bandwidth and undetermined channel fading/noise. Extensive experimental results and user study verify that our approach is favorable to yield human perception more pleasing results versus SOTA methods. Achieving the perceptually similar end-to-end image transmission quality, the proposed method can save about 50\% wireless channel bandwidth cost. Fig. \ref{Fig_introduction_demo} intuitively shows visual comparisons between our proposed method and other image coded transmission schemes. This is the first work integrating perceptual ingredient and generative method to optimize image coded transmission systems, which is enlightening for future semantic communication system design.

\emph{Notational Conventions:} Throughout this paper, lowercase letters (e.g., $x$) denote scalars, bold lowercase letters (e.g., $\mathbf{x}$) denote vectors. In some cases, $x_i$ denotes the elements of $\mathbf{x}$, which may also represent a subvector of $\mathbf{x}$ as described in the context. Bold uppercase letters (e.g., $\mathbf{X}$) denote matrices, and $\mathbf{I}_m$ denotes an $m$-dimensional identity matrix. $p_x$ denotes a probability density function (pdf) with respect to the continuous-valued random variable $x$. In addition, $\mathbb{E} [\cdot]$ denotes the statistical expectation operation, and $\mathbb{R}$ denotes the real number set. Finally, $\mathcal{N}(\mu, \sigma^2)$ denotes a Gaussian distribution with mean $\mu$ and variance $\sigma^2$.

\section{System Architecture}\label{sec_architecture}

Consider the following lossy wireless image transmission scenario. Alice is drawing a $m$-dimensional vector $\mathbf{x}$ from the source, whose probability is given as $p_{\mathbf{x}}( \mathbf{x} )$. Alice concerns how to map $\mathbf{x}$ to a $k$-dimensional vector $\mathbf{s}$, where $k$ is referred to as the \emph{channel bandwidth cost}. Then, Alice transmits $\mathbf{s}$ to Bob via a realistic wireless channel, who uses the received information $\mathbf{\hat s}$ to reconstruct an approximation to $\mathbf{x}$.

\begin{figure}[htbp]
	\setlength{\abovecaptionskip}{0.cm}
	\setlength{\belowcaptionskip}{-0.cm}
	\centering{\includegraphics[scale=0.46]{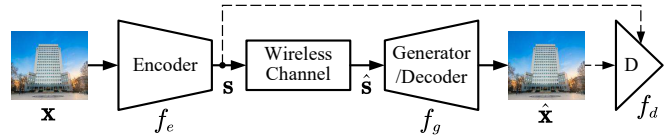}}
	\caption{Architecture of our generative deep JSCC, where the data links denoted by dashed lines are only used for model training.}\label{Fig_system_model}
	\vspace{0em}
\end{figure}

\begin{figure*}[htbp]
	\setlength{\abovecaptionskip}{0.cm}
	\setlength{\belowcaptionskip}{-0.cm}
	\centering{\includegraphics[scale=0.55]{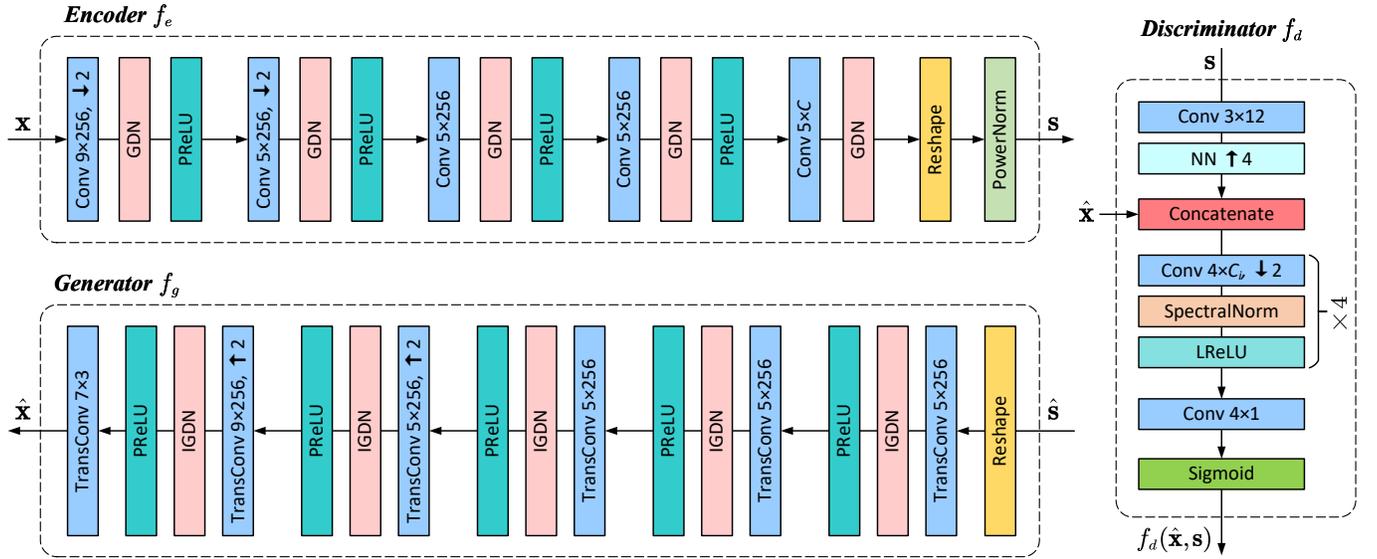}}
	\caption{Network architecture of our generative deep JSCC. Conv $k\times C$ denotes a convolutional with $C$ channels and $k\times k$ filters, and the followed $\uparrow2/\downarrow2$ indicates upscaling or downscaling with a stride of $2$. ChannelNorm refers to a normalization layer from \cite{HIFIC}. Detailed structure of residual block can also refer to \cite{HIFIC}. LReLU denotes the leaky ReLU with $\alpha=0.2$ \cite{xu2015empirical}. NN $\uparrow 4$ denotes the nearest neighbor upsampling.}\label{Fig_network_architecture}
	\vspace{0em}
\end{figure*}

The proposed generative deep JSCC framework for end-to-end wireless image transmission is illustrated in Fig. \ref{Fig_system_model}. The source image formulated as a vector $\mathbf x \in {\mathbb R}^m$ is mapped to a vector of continuous-valued channel-input symbols $\mathbf s \in {\mathbb R}^k$ via a deep neural network (DNN) based encoding function ${\mathbf s} = f_{e}( {\mathbf x})$, where $f_e$ can be parameterized as a convolutional neural network (CNN). We typically have $k<m$, and $R=k/m$ is named \emph{channel bandwidth ratio (CBR)} \cite{DJSCC} denoting the overall coding rate. Then, the analog sequence $\mathbf{s}$ is directly sent over the communication channel. The wireless channel introduces random corruptions to the transmitted symbols, denoted as a function $W( \cdot)$. The received sequence is ${\mathbf{\hat s}} = W( \mathbf{s})$, whose transition probability is ${{p_{{\mathbf{\hat s}}| {\mathbf{s}} }}( {{\mathbf{\hat s}}| \mathbf{s} } )}$. In this paper, we exemplarily consider the most widely used additive white Gaussian noise (AWGN) channel model such that the transfer function is ${\mathbf{\hat s}} = W( \mathbf{s} ) = \mathbf{s} + \mathbf{n}$ where each component of the noise vector $\mathbf{n}$ is independently sampled from a Gaussian distribution, i.e., $\mathbf{n} \sim \mathcal{N}(0, {\sigma^2}{\mathbf{I}}_k)$, where ${\sigma^2}$ is the average noise power. Other channel models can also be similarly incorporated by changing the channel transition function.

The receiver also comprises a generator/decoder function ${{\mathbf {\hat x}}} = f_{g}( {{\mathbf {\hat s}}} )$ to recover the corrupted signal ${\mathbf{\hat s}}$ as ${{\mathbf {\hat x}}}$, where $f_g$ can also be a format of CNN \cite{DJSCC}. In contrast to traditional deep JSCC, the target of $f_g$ is not only recovering the transmitted image $\mathbf{x}$ to a certain degree, but also trying to generate ${{\mathbf {\hat x}}}$ that is consistent with the real conditional distribution $p_{\mathbf{x}|\mathbf{s}}$. The latter is contributed by the adversarial loss given by the discriminator $f_d$ \cite{goodfellow2014generative} as shown in Fig. \ref{Fig_system_model}, this idea comes from conditional generative adversarial networks (cGANs) \cite{mirza2014conditional}. Therefore, the objective for our generative deep JSCC can be expressed as
\begin{equation}\label{eq_overall_target}
\begin{aligned}
  \mathop {\min }\limits_{{f_e},{f_g}} \mathop {\max }\limits_{{f_d}} & \Big( \underbrace {\mathbb{E}\big[-\log{f_d(\mathbf{x},\mathbf{{s}})} \big] + \mathbb{E}\big[-\log({1 - f_d(f_g(\mathbf{\hat{s}}),\mathbf{{s}})}) \big]}_{\text{GAN loss}} \\
  ~ & + \beta  \underbrace {\mathbb{E}\big[ d(\mathbf{x}, f_g(\mathbf{\hat{s}})) \big]}_{\text{distortion}} + \lambda R \Big),
\end{aligned}
\end{equation}
where $\lambda$ balances the trade-off between the CBR against the distortion and GAN loss terms, and $\beta$ balances between the two types of losses. Herein, in deep JSCC models, the CBR term $R$ is predetermined by the input/output dimensions of the encoder network. Even though it is more reasonable to estimate the entropy of codeword sequence as \cite{dai2021nonlinear} for further saving the bandwidth cost, it also requires explicit entropy-modeled rate as a loss term. We leave the exploration of this issue for future work. In this paper, we also follow the standard model of deep JSCC to manually set $R$, in this simplified setting, the CBR term in \eqref{eq_overall_target} is indeed a fixed term.

Given the CBR $R$, our generative deep JSCC model training is divided into two phases in each iteration. The goal is to create a pair of encoder $f_e$ and generator $f_g$ to ``fool'' the discriminator $f_d$ into believing the reconstructed image samples $\mathbf{\hat{x}}$ are real while also considering the influence of an imperfect wireless channel. The loss functions are written as follows.
\begin{itemize}
  \item Training the encoder $f_e$ and generator $f_g$ with the loss function $L_{E,G}$:
  \begin{equation}\label{eq_generative_loss}
  \begin{aligned}
    & \mathop {\min }\limits_{{f_e},{f_g}}L_{E,G} = \mathop {\min }\limits_{{f_e},{f_g}} \Big( {\beta_{d}}\underbrace{\mathbb{E}\big[-\log({f_d(f_g(\mathbf{\hat{s}}),\mathbf{{s}})}) \big]}_{\text{generative loss}} + \\
    & {\beta_p}\underbrace{\mathbb{E}\big[ d_{\text{LPIPS}}(\mathbf{x}, f_g(\mathbf{\hat{s}})) \big]}_{\text{perceptual loss}} + {\beta_m}\underbrace{\mathbb{E}\big[ d_{\text{MSE}}(\mathbf{x}, f_g(\mathbf{\hat{s}})) \big]}_{\text{regular MSE loss}}  \Big).
    \end{aligned}
  \end{equation}
\end{itemize}
The discriminator function $f_d$ in \eqref{eq_generative_loss} includes two inputs which obey the format of cGANs \cite{mirza2014conditional}. Each data sample $\mathbf{\hat{x}} = f_g(\mathbf{\hat{s}})$ is associated with the channel corrupted codeword $\mathbf{\hat{s}}$ as the condition. The generator $f_g$ is trained to map the received samples $\mathbf{\hat{s}}$ from a fixed distribution $p_{\mathbf{\hat{s}}|\mathbf{x}}$ to the real distribution $p_{\mathbf{{x}}|\mathbf{{s}}}$, where $p_{\mathbf{\hat{s}}|\mathbf{x}}$ is determined by the encoder function $f_e$ and channel transition function $W$. $d_{\text{LPIPS}}$ denotes the perceptual LPIPS loss function \cite{lpips}. As the learning-based LPIPS metric has many different configurations, we exemplarily adopt the Alexnet-based \cite{alexnet} LPIPS loss to help model training in a perception-aware way. $d_{\text{MSE}}$ denotes the regular MSE distortion function. ${\beta_d}$, ${\beta_p}$, ${\beta_m}$ control the trade-off between three distortion terms.

\begin{itemize}
  \item Training the discriminator $f_d$ with the loss function $L_D$:
  \begin{equation}\label{eq_discriminative_loss}
  \begin{aligned}
    & \mathop {\max }\limits_{{f_d}}{L_{D}} = \mathop {\max }\limits_{{f_d}} \Big( \mathbb{E}\big[-\log({1-f_d(f_g(\mathbf{\hat{s}}),\mathbf{{s}})}) \big]\\
    &  + \mathbb{E}\big[-\log({f_d(\mathbf{x},\mathbf{{s}})}) \big] \Big).
  \end{aligned}
  \end{equation}
\end{itemize}
The channel corrupted sequence $\mathbf{\hat{s}}$ is associated with the source image sample $\mathbf{x}$ in \eqref{eq_discriminative_loss}, i.e., $\mathbf{\hat{s}} = W(f_e(\mathbf{x}))$. The discriminator $f_d$ is mapping the input tuple $(\mathbf{x},\mathbf{{s}})$ to the probability that it is a sample from the real distribution $p_{\mathbf{{x}}|\mathbf{{s}}}$ rather than the fake sample generated by $f_g$.

\section{Network Architecture}\label{sec_implementation}

\begin{figure*}[t]
\setlength{\abovecaptionskip}{0.cm}
\setlength{\belowcaptionskip}{-0.cm}
\begin{center}
		\hspace{-.10in}
		\subfigure[]{	
            \includegraphics[scale=0.23]{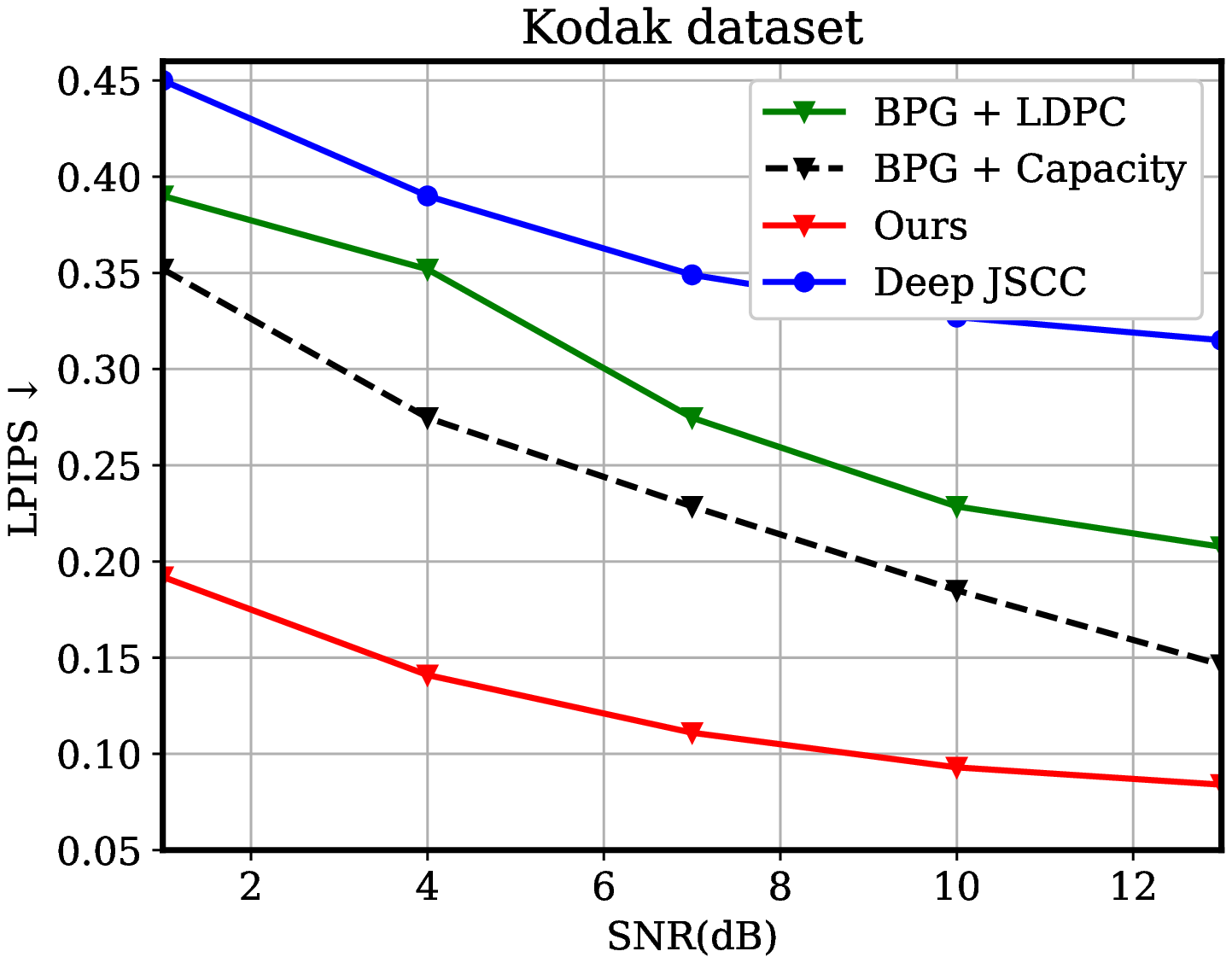}
		}
		\hspace{-.3in}
		\quad
		\subfigure[]{
			\includegraphics[scale=0.23]{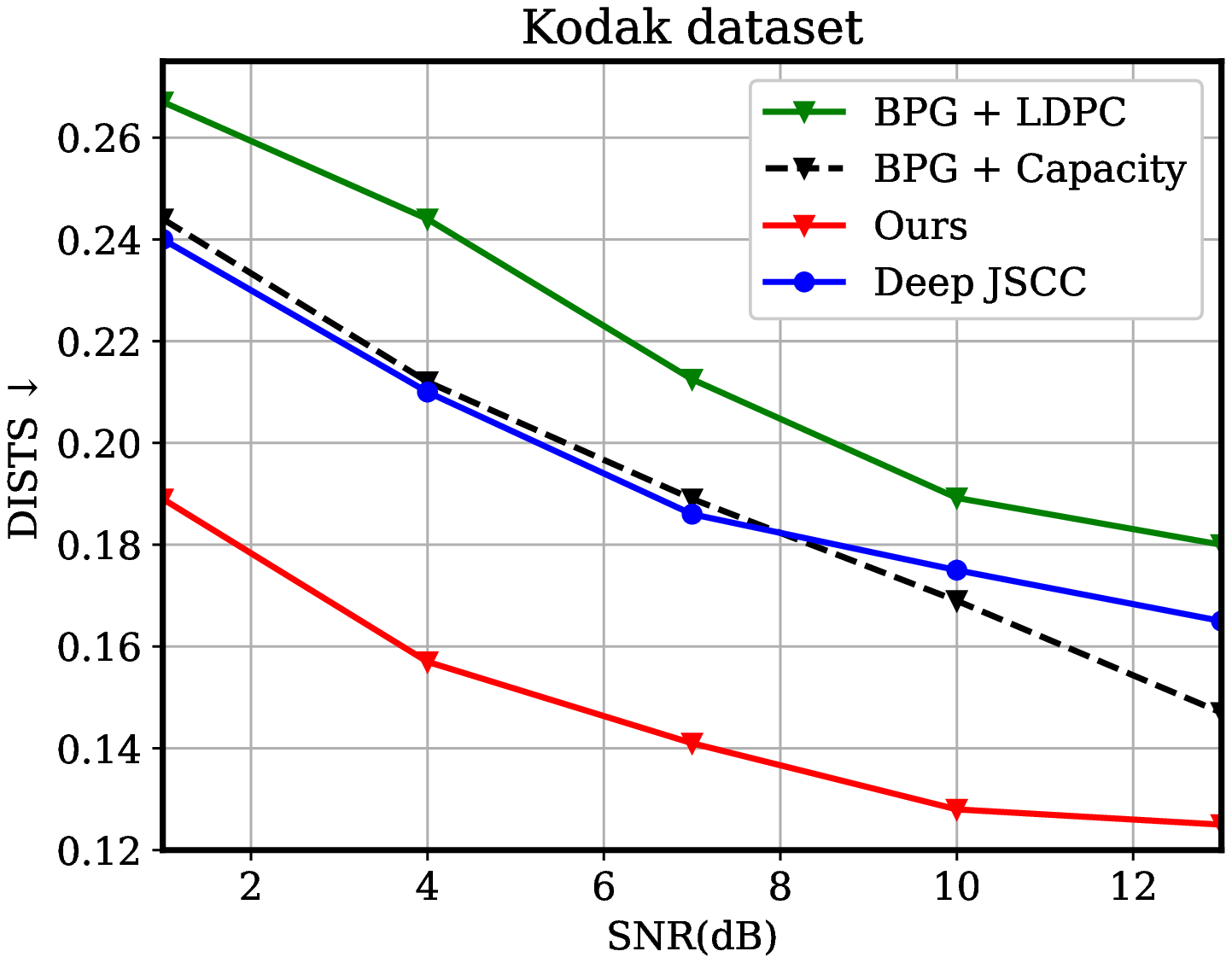}
		}
		\hspace{-.3in}
		\quad
		\subfigure[]{
			\includegraphics[scale=0.23]{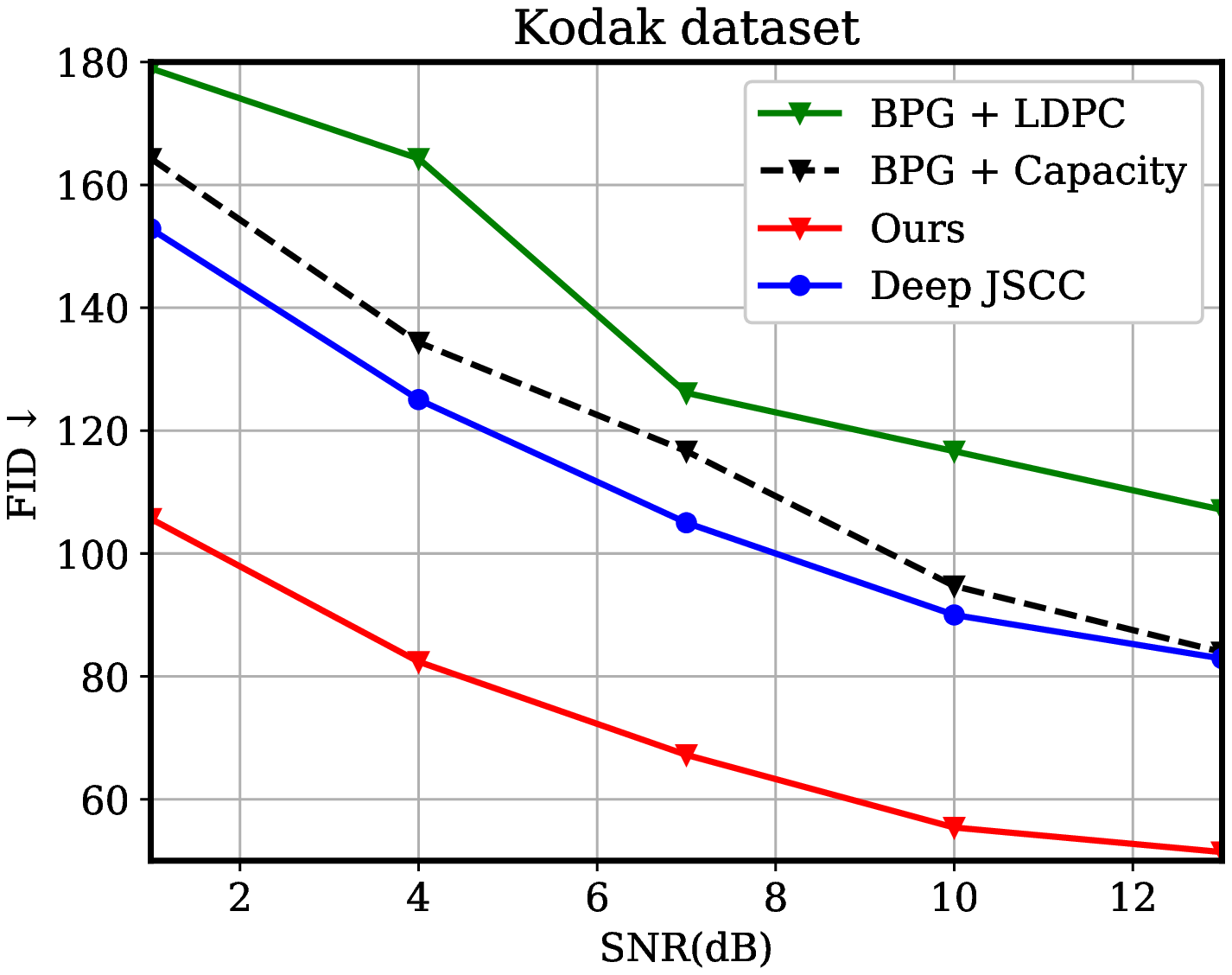}
		}
		\hspace{-.3in}
		\quad
		\subfigure[]{
			\includegraphics[scale=0.23]{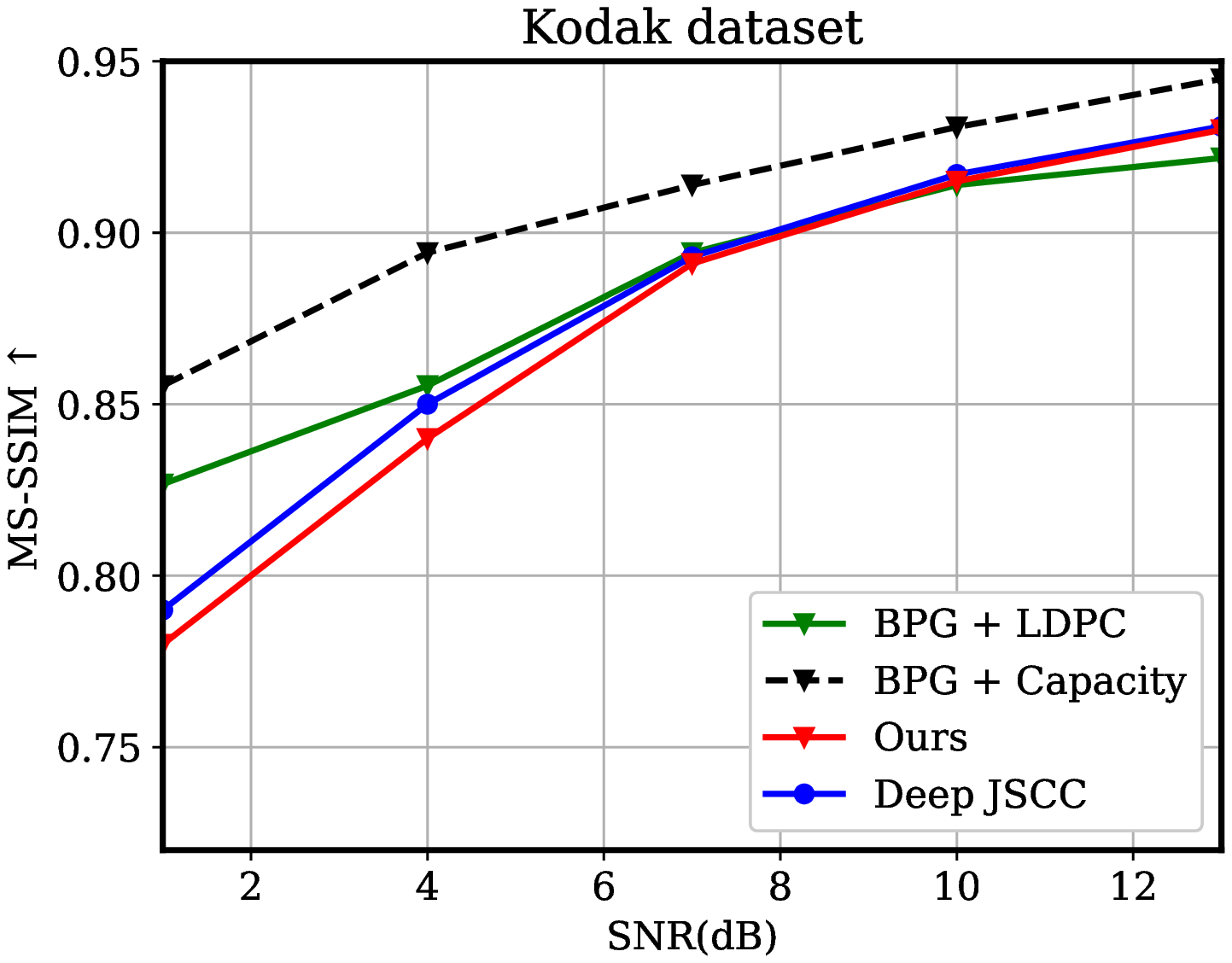}
		}
        \hspace{-.3in}
		\quad
		\subfigure[]{
			\includegraphics[scale=0.23]{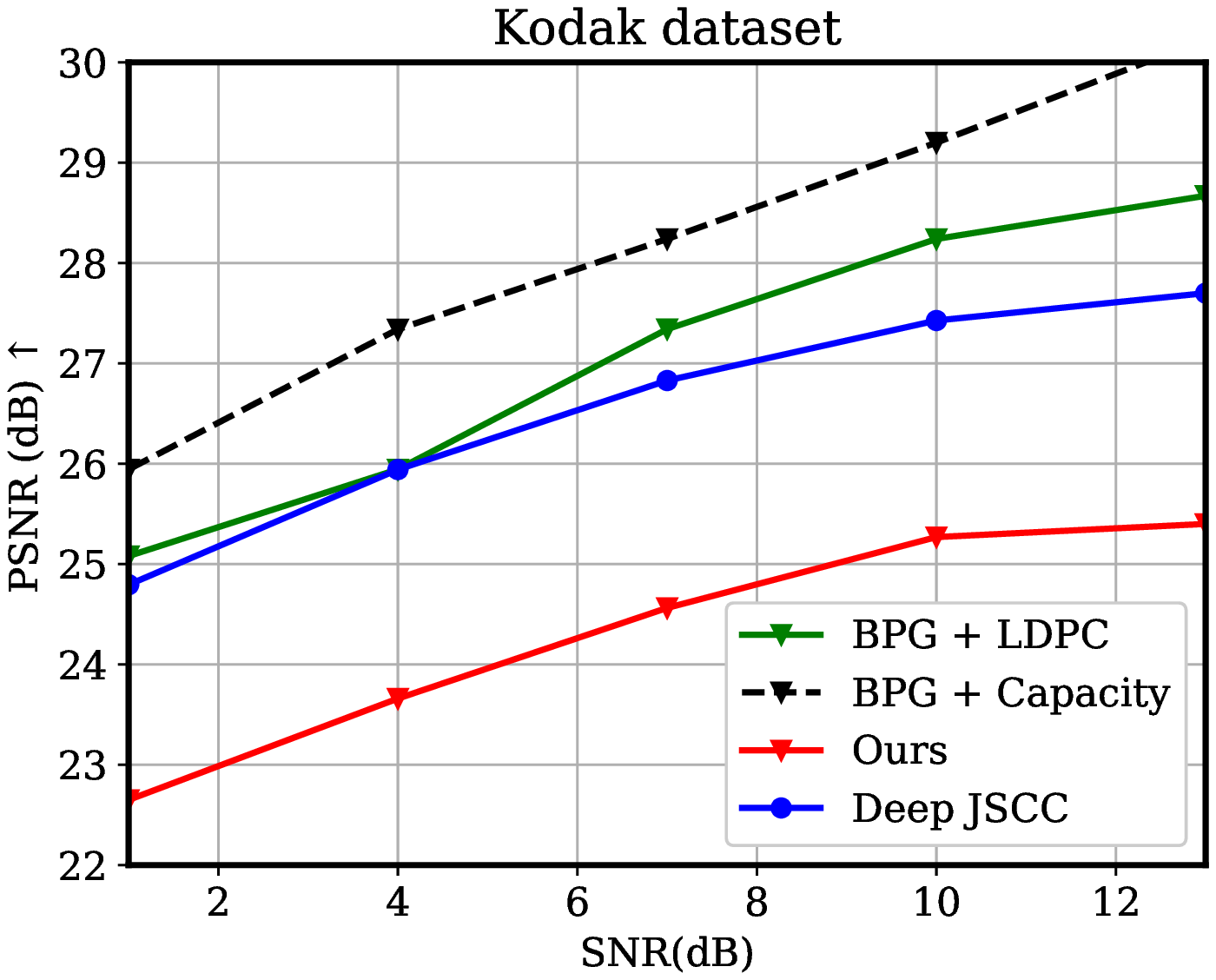}
		}

        \hspace{-.10in}
		\subfigure[]{	
            \includegraphics[scale=0.23]{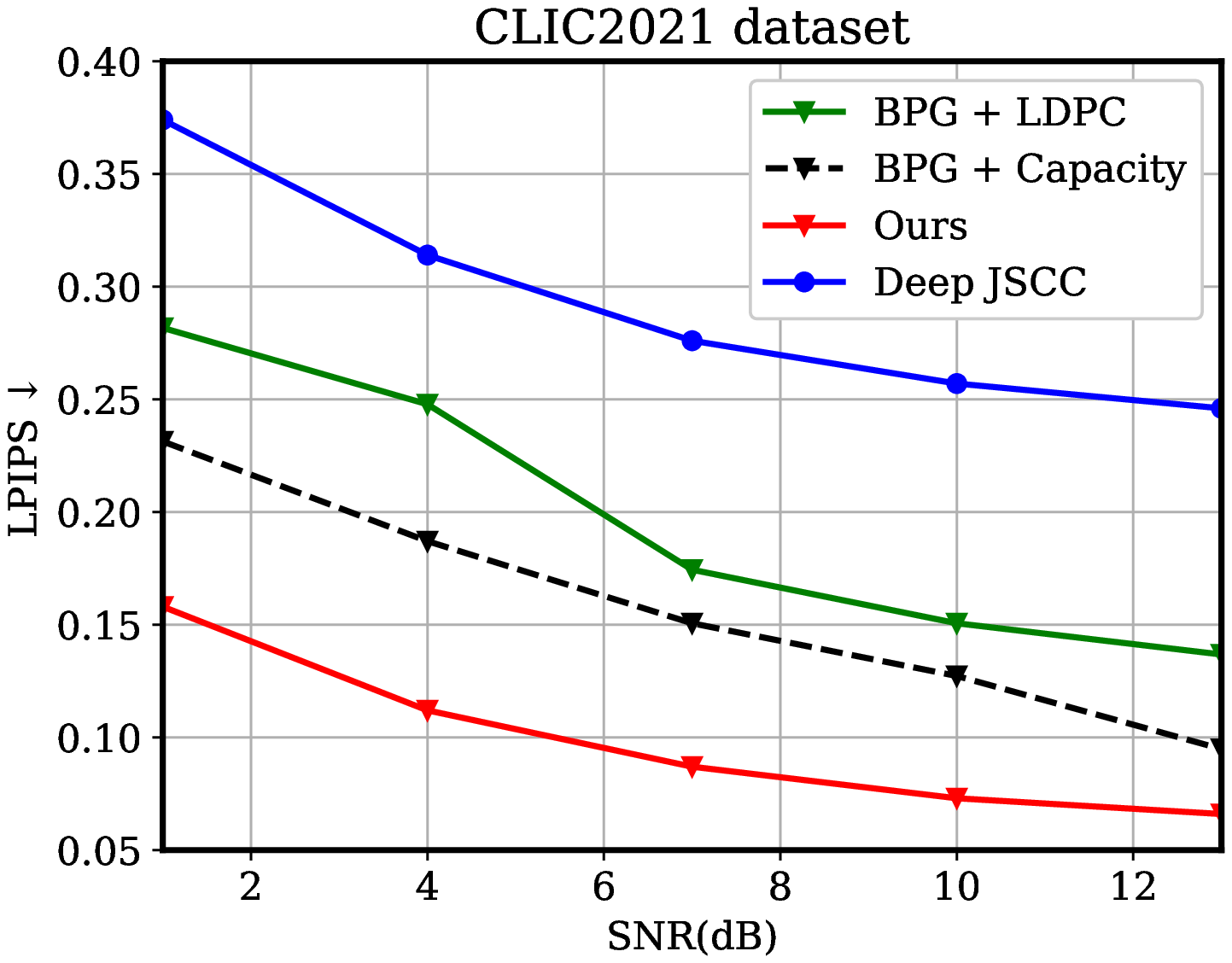}
		}
		\hspace{-.3in}
		\quad
		\subfigure[]{
			\includegraphics[scale=0.23]{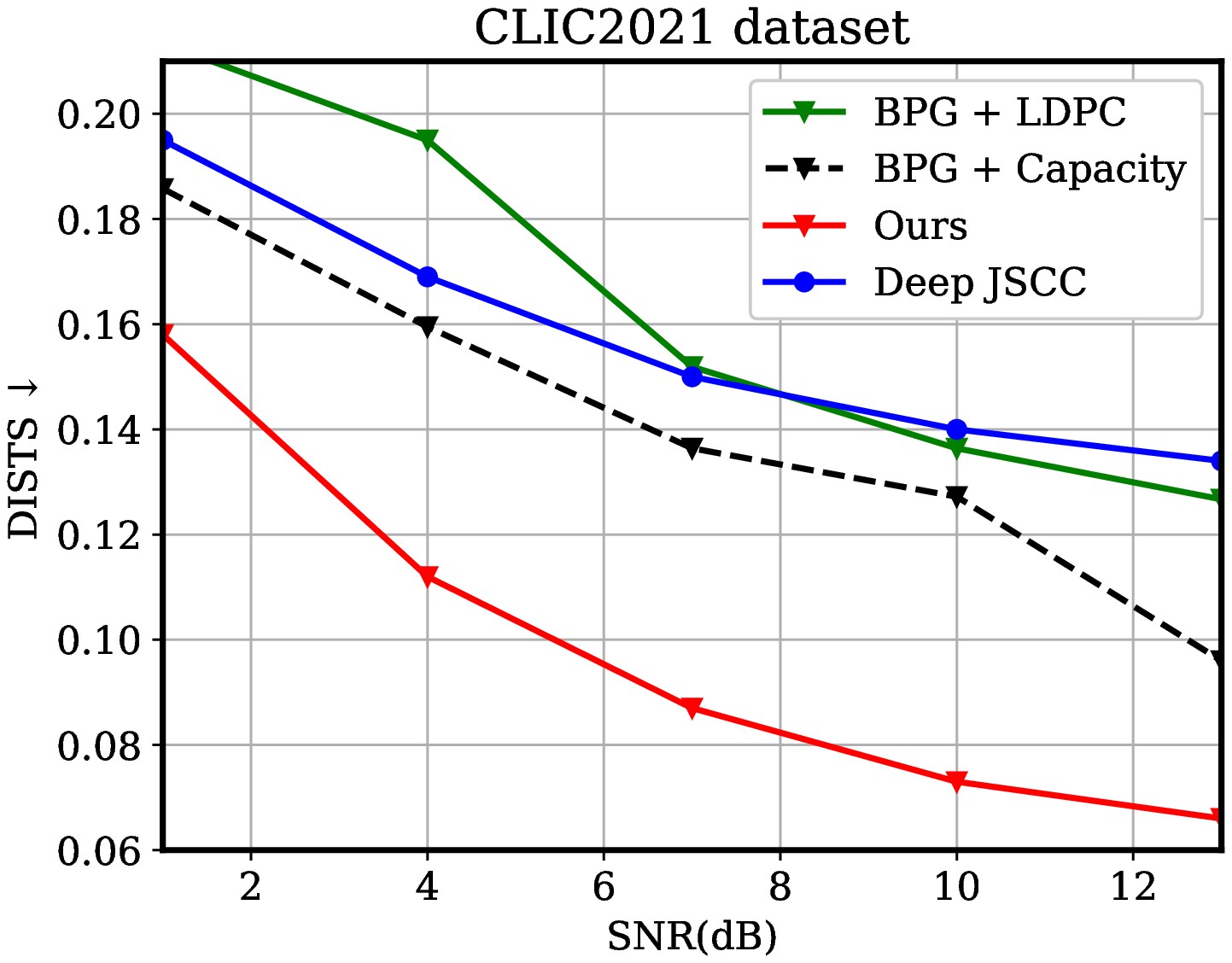}
		}
		\hspace{-.3in}
		\quad
		\subfigure[]{
			\includegraphics[scale=0.23]{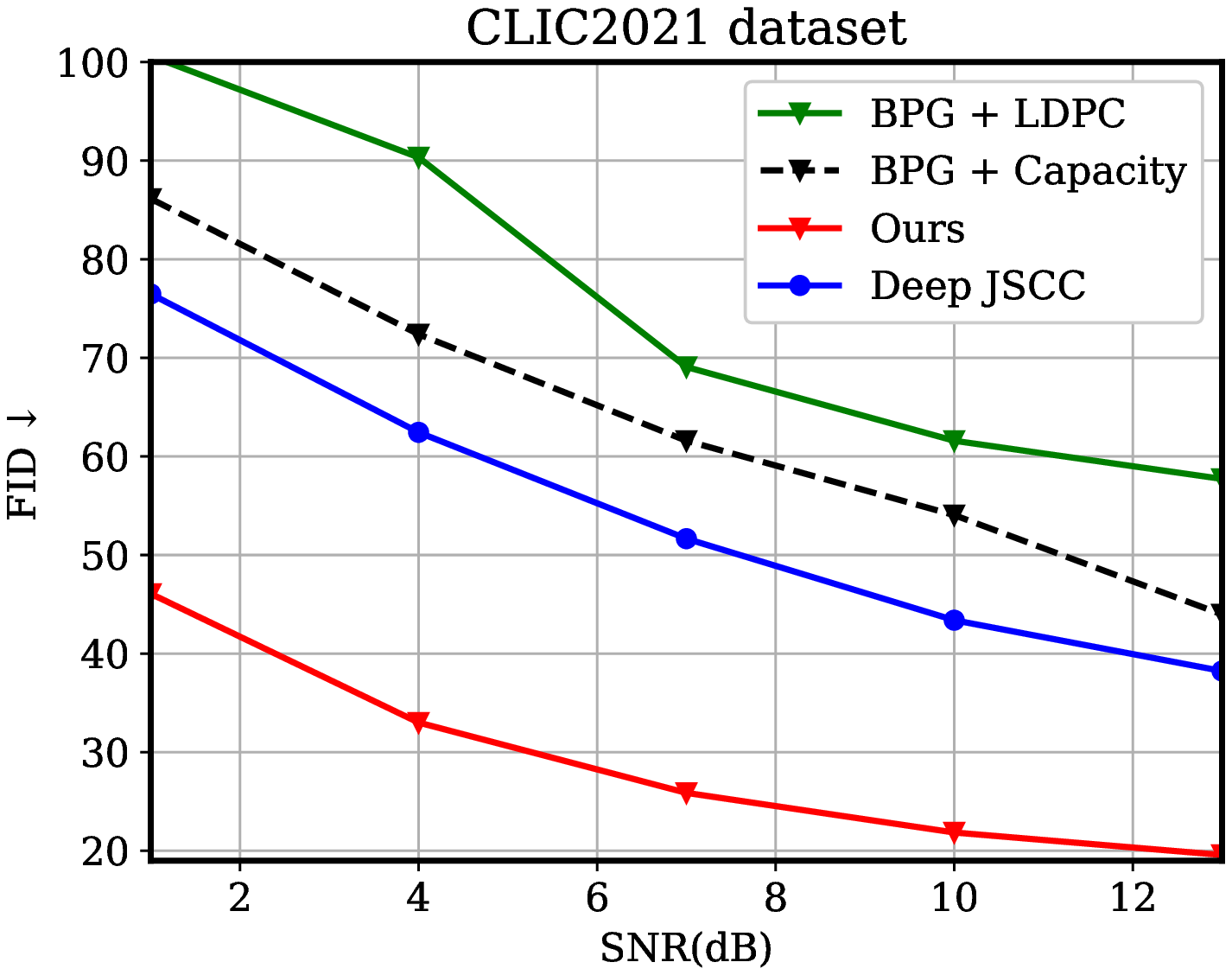}
		}
		\hspace{-.3in}
		\quad
		\subfigure[]{
			\includegraphics[scale=0.23]{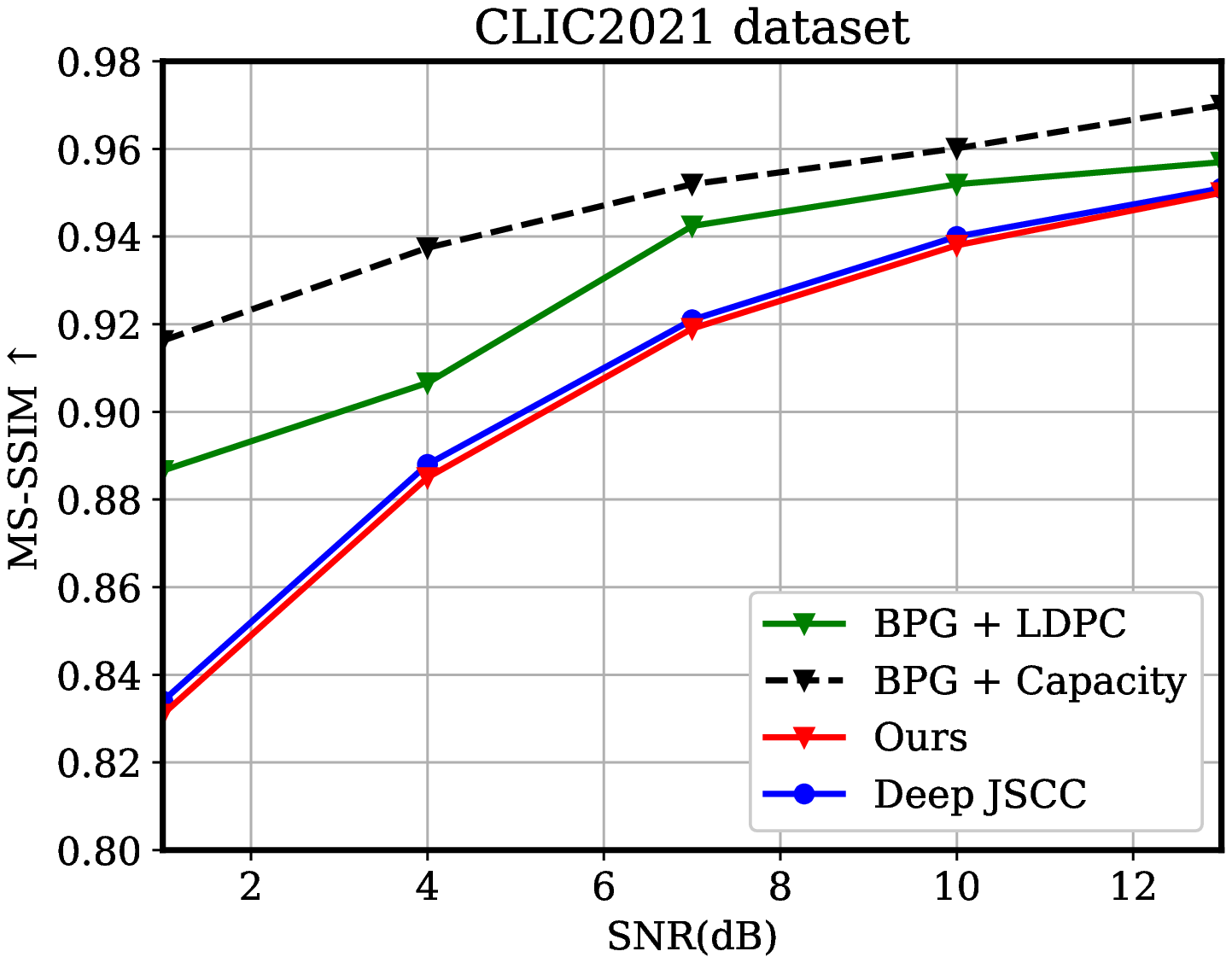}
		}
        \hspace{-.3in}
		\quad
		\subfigure[]{
			\includegraphics[scale=0.23]{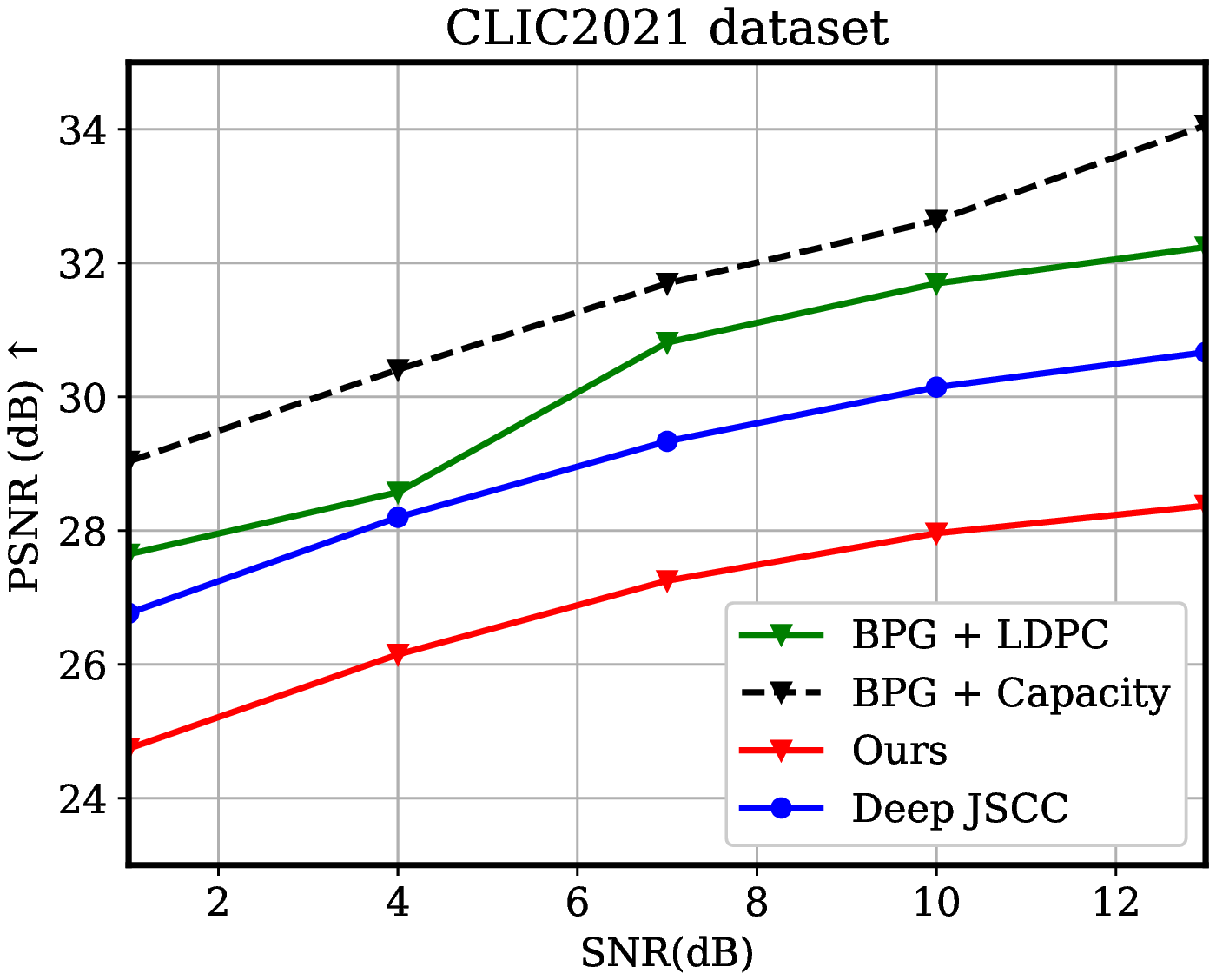}
		}
		\caption{SNR-perception and SNR-distortion curves on Kodak and CLIC2021 datasets. Arrows in the vertical axis indicate whether lower is better ($\downarrow$), or higher is better ($\uparrow$). The experiments are taken over the AWGN channel with varying channel SNRs with the channel bandwidth ratio (CBR) = 1/48.}
        \label{Fig_SNR results}
\end{center}
\vspace{-0.5em}
\end{figure*}

\begin{figure*}[htbp]
\setlength{\abovecaptionskip}{0.cm}
\setlength{\belowcaptionskip}{-0.cm}
\begin{center}
		\hspace{-.10in}
		\subfigure[]{	
            \includegraphics[scale=0.24]{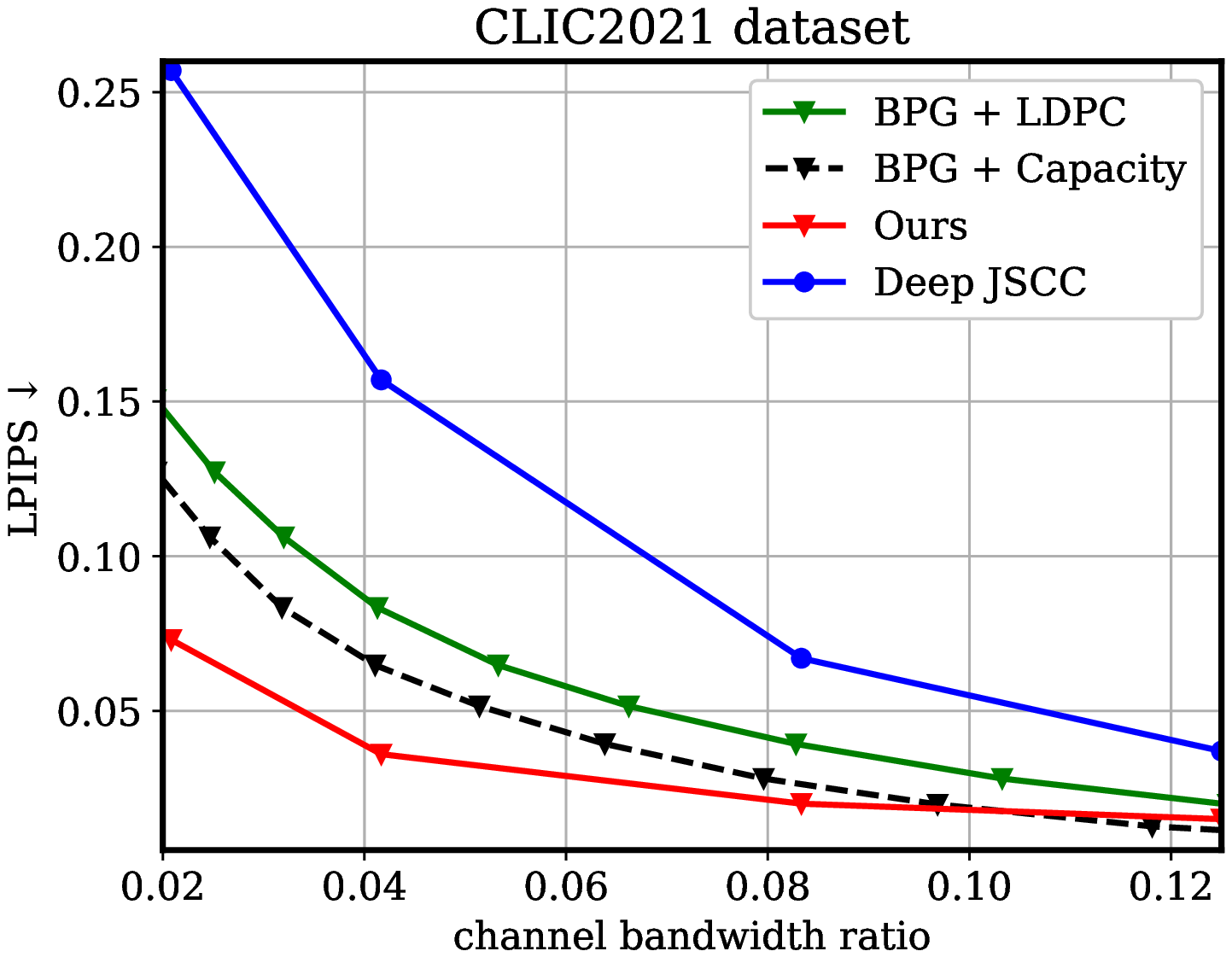}
		}
		\hspace{-.3in}
		\quad
		\subfigure[]{
			\includegraphics[scale=0.24]{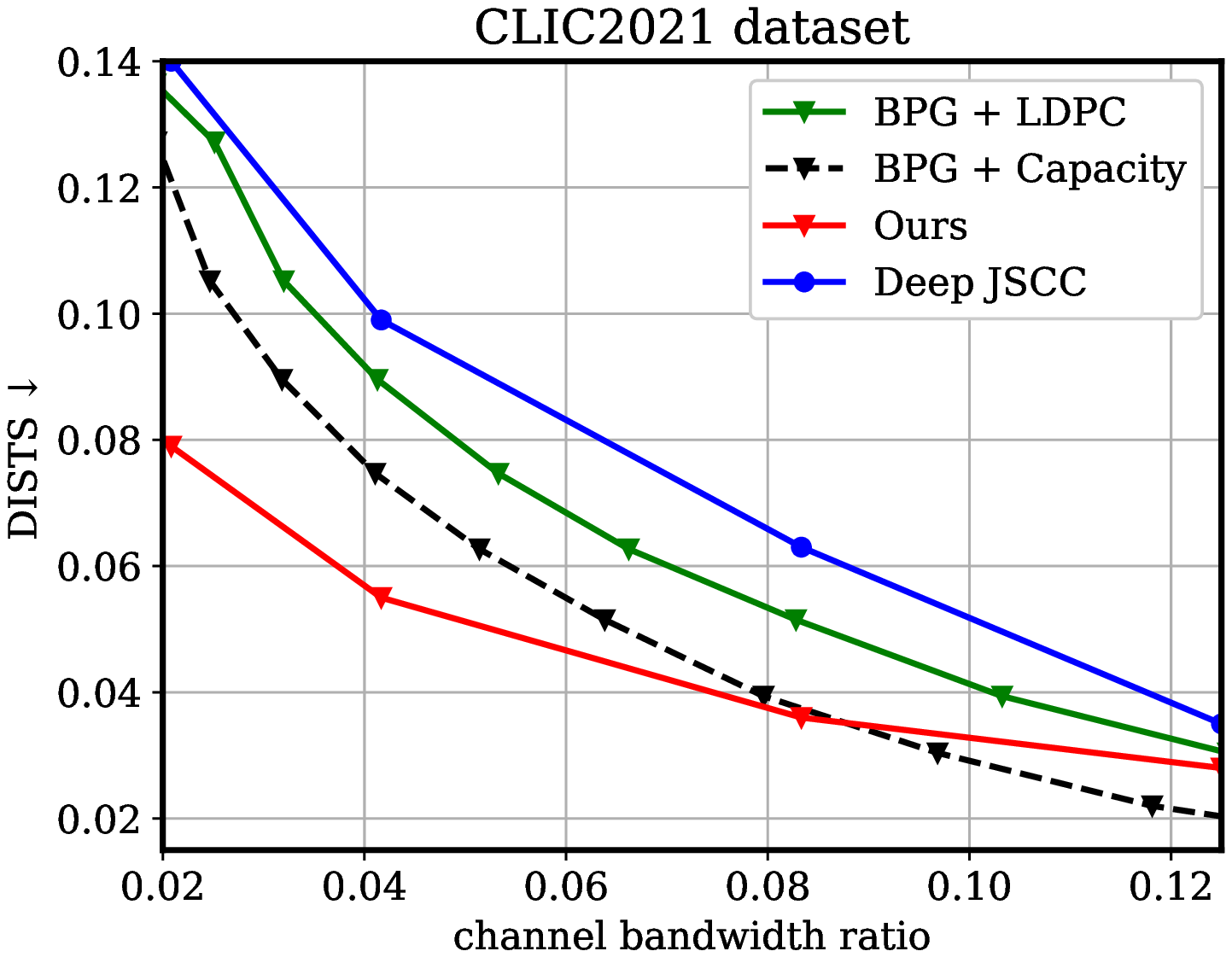}
		}
		\hspace{-.3in}
		\quad
		\subfigure[]{
			\includegraphics[scale=0.24]{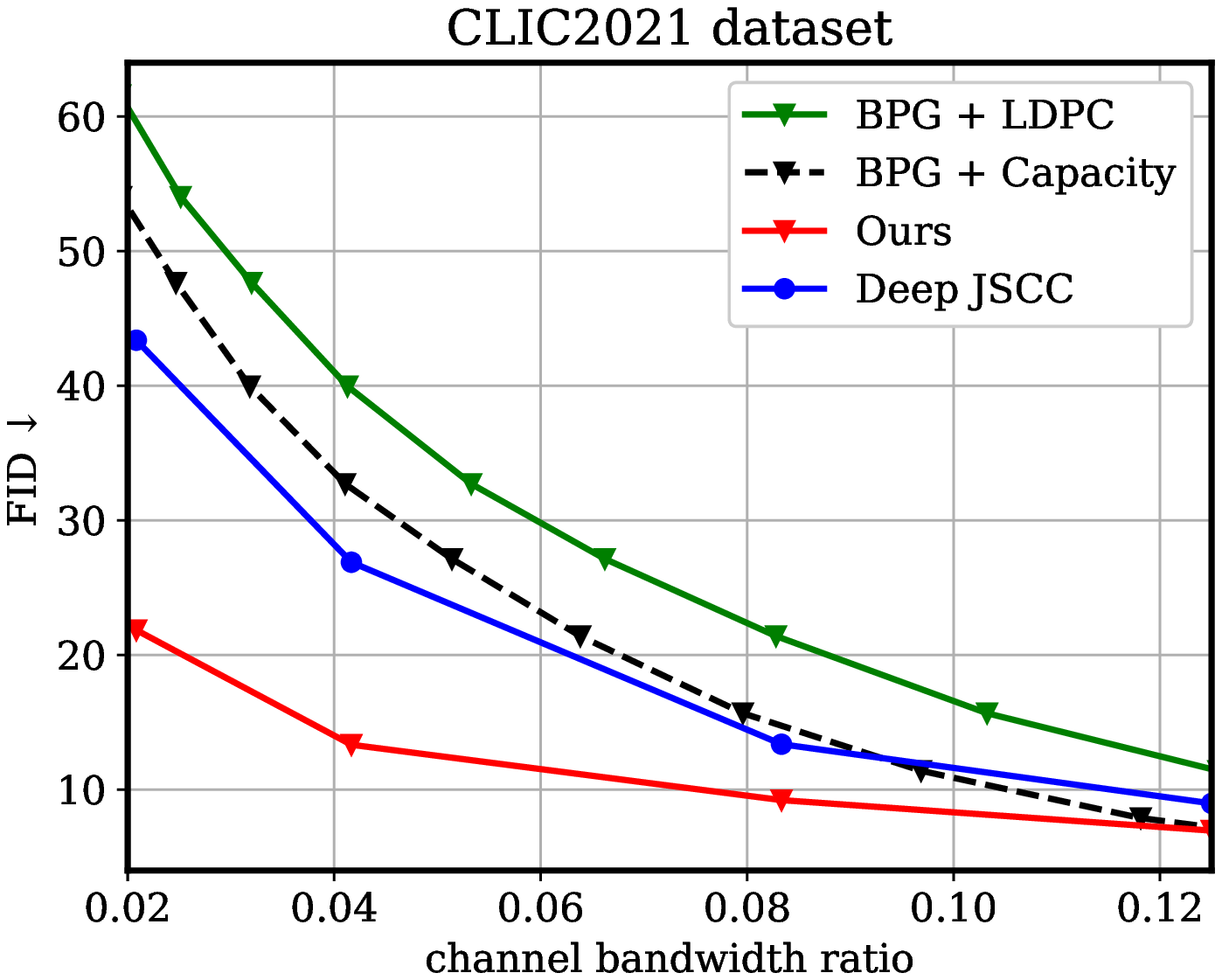}
		}
		\hspace{-.3in}
		\quad
		\subfigure[]{
			\includegraphics[scale=0.24]{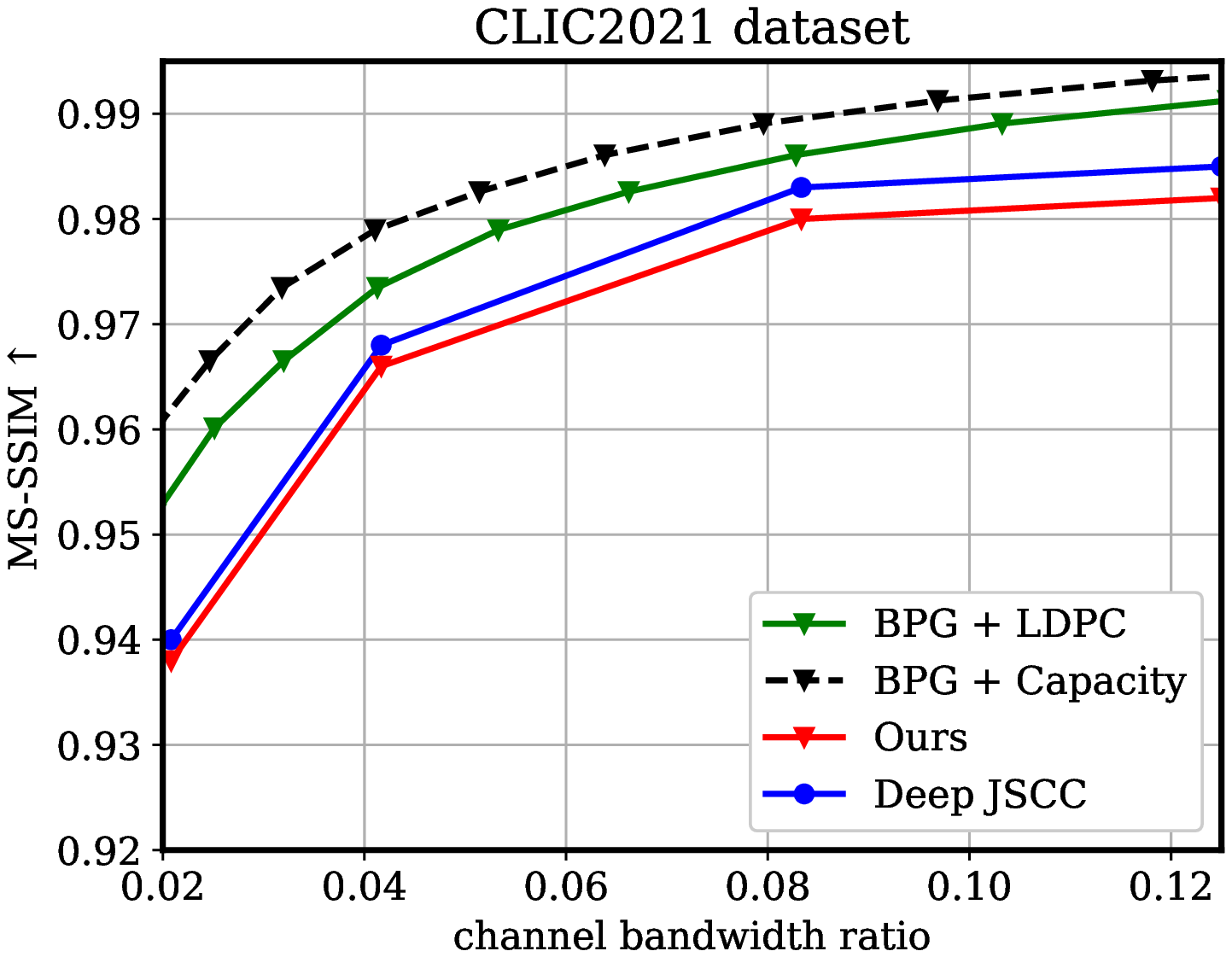}
		}
        \hspace{-.3in}
		\quad
		\subfigure[]{
			\includegraphics[scale=0.24]{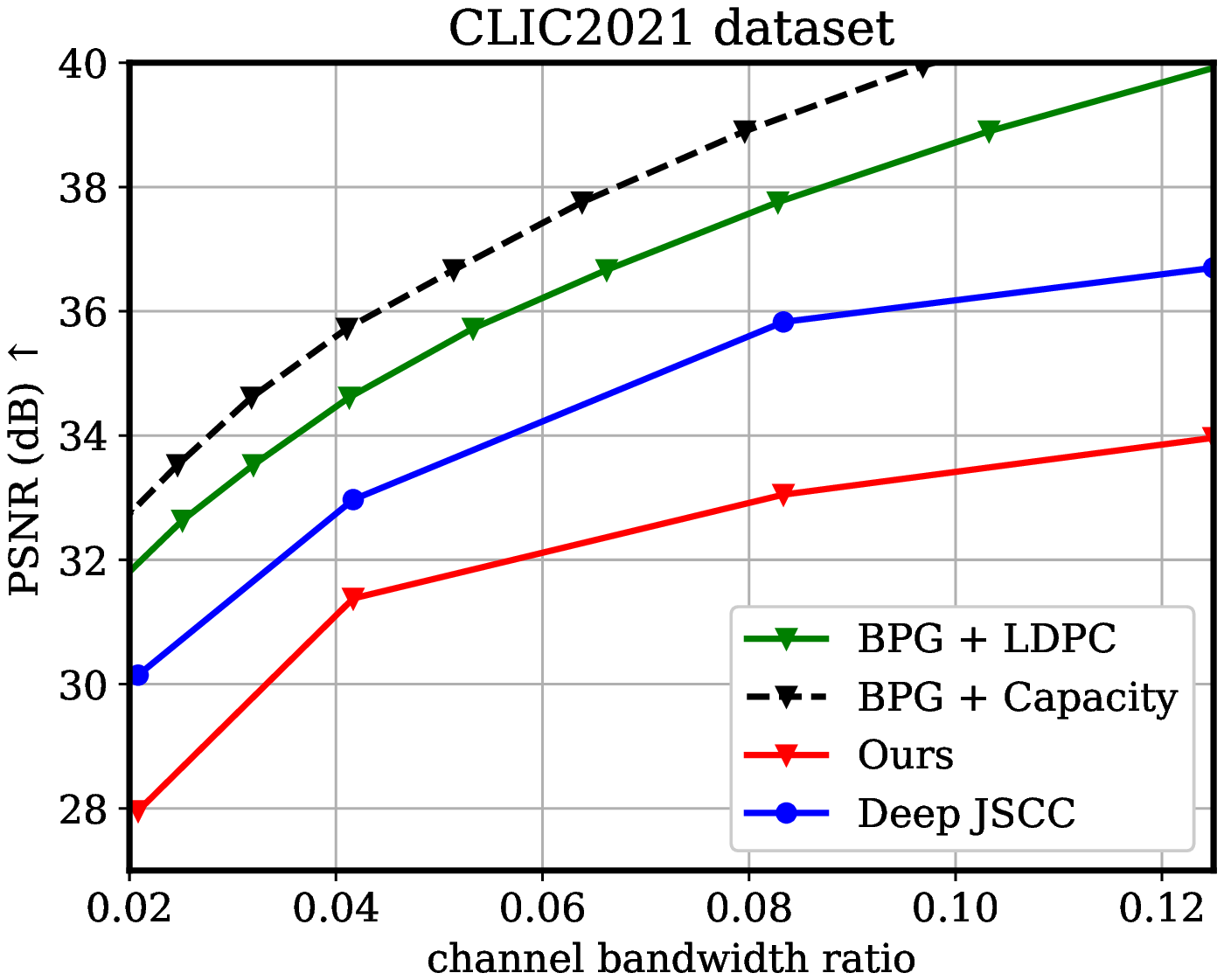}
		}

		\caption{CBR-distortion and CBR-perception curves on CLIC2021. Arrows in the vertical axis indicate whether lower is better ($\downarrow$), or higher is better ($\uparrow$). The experiments are taken over the AWGN channel at SNR = 10dB.}

        \label{Fig_CBR results}
\end{center}
\vspace{-1.5em}
\end{figure*}

Our network architecture is shown in Fig. \ref{Fig_network_architecture}, including the encoder $f_e$, generator $f_g$, and discriminator $f_d$. The encoder $f_e$ uses convolution neural networks (CNNs) to extract features of the source image $\mathbf{x}$ and reshape them to power normalized channel-input symbols $\mathbf{s}$. The decoder $f_g$ performs the inverse operations of the encoder.

Our discriminator network $f_d$ is basically the same as that in PatchGAN \cite{li2016precomputed}, which penalizes structure at the scale of image patches, i.e., it tries to identify if each patch of the reconstruction is real or fake. The patch size is set to $16 \times 16$ pixels in our implementation. Besides, we also make use of the structure of cGAN \cite{mirza2014conditional} by concatenating the reconstructed image with an upscaled deep JSCC codeword to provide the contextual information. In particular, four cascaded modules of the form ``convolution-SpectrualNorm-LeackyReLU'' \cite{miyato2018spectral} is used to downsample the concatenated feature map progressively, and the number of kernels in convolutions $C_i$ is 64, 128, 256, and 512, respectively. After the last layer, a convolution with the Sigmoid activation indicates the discriminator's true or false judgments for each patch.

During the model training, we firstly pretrain the encoder $f_e$ and generator $f_g$ only using the perceptual loss $d_{\text{LPIPS}}$ and the regular MSE loss $d_{\text{MSE}}$. The function of the pretraining step is to accelerate the experiments and makes them more stable. In the second step, we freeze the encoder $f_e$ and generator $f_g$ and only optimize the discriminator $f_d$, which enables it to penalize structure at image patches produced by pretrained weights. Finally, we adopt an iterative approach to alternately training the generative network ($f_e$ and $f_g$) and the discriminator network ($f_d$).

\section{Experimental Results}\label{sec_results}

\subsection{Experimental setup}

\subsubsection{Training Details}

The training set consists of 720,000 images sampled from the Open Images dataset \cite{OpenImage}. During training, we form a batch of 12 patches by randomly cropping a $256 \times 256$ patch from each image. We first pretrain the network except for the discriminator $f_d$ by about 100,000 iterations jointly using the LPIPS perceptual loss and the regular MSE loss with $\beta_p = 1.0$ and $\beta_m = 10^{-5}$ under a uniform distribution of $\text{SNR}_{\text{train}}$ $\in{ \{1, 4, 7, 10, 13\}}$ dB. Then, we freeze the encoder and generator and optimize the discriminator for about 10,000 iterations. Finally, we alternately train the generative network ($f_e$ and $f_g$) by setting $\beta_g = 10^{-3}$. We use the Adam optimizer with a learning rate of $10^{-4}$.

\subsubsection{Evaluation Protocols}

We evaluate our method on the widely-used Kodak dataset \cite{Kodak} (24 images, $768 \times 512 $ pixels) and the test set of the CLIC2021 dataset \cite{CLIC2021} (60 images, up to $2048 \times 1890$ pixels). The comparison schemes include the deep JSCC optimized for MSE (marked as ``deep JSCC (MSE)'') and the classical coded transmission schemes (BPG + LDPC). In particular, for the deep JSCC scheme, we adopt the \emph{Deep JSCC (no feedback)} scheme in \cite{DJSCCF}, which achieves better performance than the original paper \cite{DJSCC}. For traditional coded transmission schemes, we employ the BPG \cite{BPG} image codec for source coding and LDPC followed by quadrature amplitude modulation (QAM) for channel coding and modulation. We further adopt BPG combined with an ideal capacity-achieving channel code, which can be viewed as the performance upper bound of the traditional coded transmission scheme. For brevity, ``BPG + LDPC'' and ``BPG + Capacity'' refers to the two comparison schemes, respectively.

\subsubsection{Metrics}  

We evaluate our models and the comparison schemes in classical metric PSNR and MS-SSIM \cite{msssim} as well as the emerging deep learning based perceptual metric LPIPS \cite{lpips}, DISTS \cite{ding2020image}, FID \cite{heusel2017gans}. PSNR corresponds to the regular MSE distortion in \eqref{eq_generative_loss}. MS-SSIM is the most widely used perceptual metric despite failing to account for many nuances of human perception \cite{lpips}. LPIPS and DISTS both can predict well with human perceptual judgment by computing the distance in the feature space of the image classification network. FID is widely used to assess the quality and diversity of generated images especially from GAN, which measures the similarity between the distribution of reference images and generated images in the feature space of an Inception network. As the image resolutions vary in a large range, we use the non-overlapping 2155 patches size of 256 pixels for the CLIC2021 dataset and 144 same size patches for the Kodak dataset.

Inspired by \cite{agustsson2019}, we also carry out a user study to more holistically evaluate the visual quality of our results. Considering the tiny distortion between the original image $\mathbf{x}$ and the reconstructed image $\hat{\mathbf{x}}$ at the high CBR, we design the user study for two models with CBR = 1/48 and CBR = 1/24 and BPG at CBR ranging from 0.029 to 0.058 at SNR = 10dB. We crop the reconstructed images on the Kodak dataset to $256 \times 256 $ patches which are respectively produced from our models and BPG + LDPC. Then we compose a questionnaire comprising of 46 reconstructed patch pairs. Each patch pair include two patches produced by different methods but from the same patch. 20 random participants are asked to choose which patch corresponds to human visual perception. We report the mean percentage of preference for each choice.

\subsection{Results Analysis}

\subsubsection{Quantitative Comparison}

Fig. \ref{Fig_SNR results} shows SNR-distortion and SNR-perception results at the CBR = 1/48 on the Kodak dataset and the CLIC2021 testset. Solid red line and solid green line plot the performance of our model and deep JSCC, where the training SNR value is consistent with the testing SNR value. For the BPG + LDPC scheme, we present the envelope of the best performing configurations at each SNR among different combinations of LDPC coding rate and modulations. Moreover, as shown in Fig. \ref{Fig_CBR results}, we further present the CBR-perception and CBR-distortion curves over AWGN channel at $\text{SNR} = 10$dB on CLIC2021. For the BPG + LDPC scheme, following \cite{DJSCCF}, we use a 2/3 rate (4096, 6144) LDPC code with 16-ary QAM to ensure reliable transmission and adjust the BPG to the largest compression rate within the CBR constraint.

Results indicate that our model dominates in terms of all three perceptual quality indices (LPIPS, DISTS, and FID). In particular, our scheme can achieve comparable or even better perceptual quality values than other approaches while saving 2x to 4x bandwidth cost in the same SNR or otherwise achieving 4dB to 8dB SNR gain using the same CBR. It validates the effectiveness of the proposed compound loss function in \eqref{eq_overall_target}. Even though perceptual metrics match better with human judgments \cite{ding2021comparison}, more convincing results can be made through subjective testing in the following user study. Besides, we also note that our scheme presents some performance degradation in the PSNR metric compared to the MSE optimized deep JSCC scheme. These verify that our scheme prefers to generate clear textures instead of a blur, resulting in lower pixel-wise metrics than comparison schemes. Performance degradation becomes negligible when considering another objective metric MS-SSIM, which is more correlated with perceptual quality than PSNR.

\subsubsection{User Study}

\begin{figure}[t]
	\setlength{\abovecaptionskip}{0.cm}
	\setlength{\belowcaptionskip}{-0.cm}
	
	\centering{\includegraphics[scale=0.39]{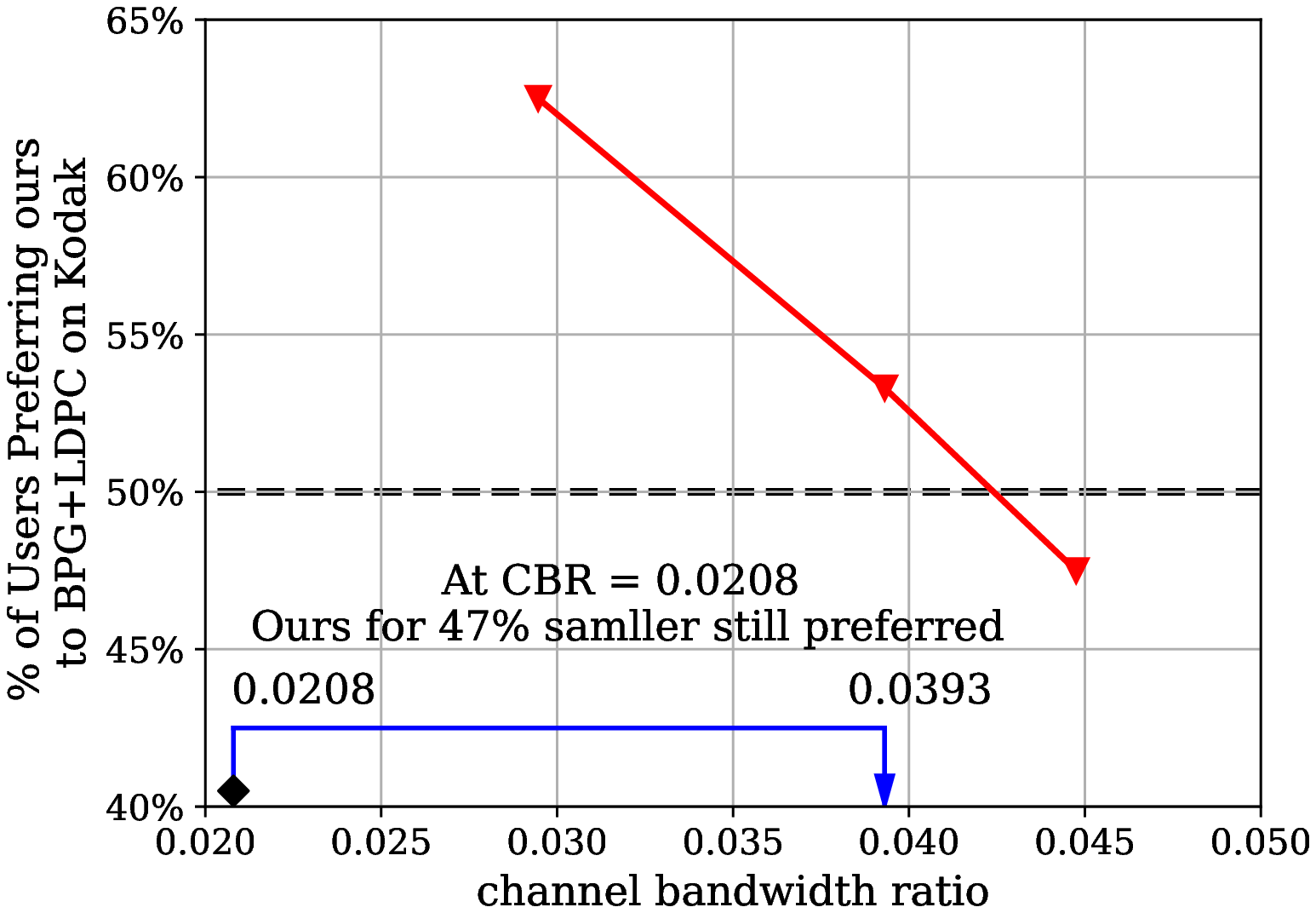}}
	\centering{\includegraphics[scale=0.39]{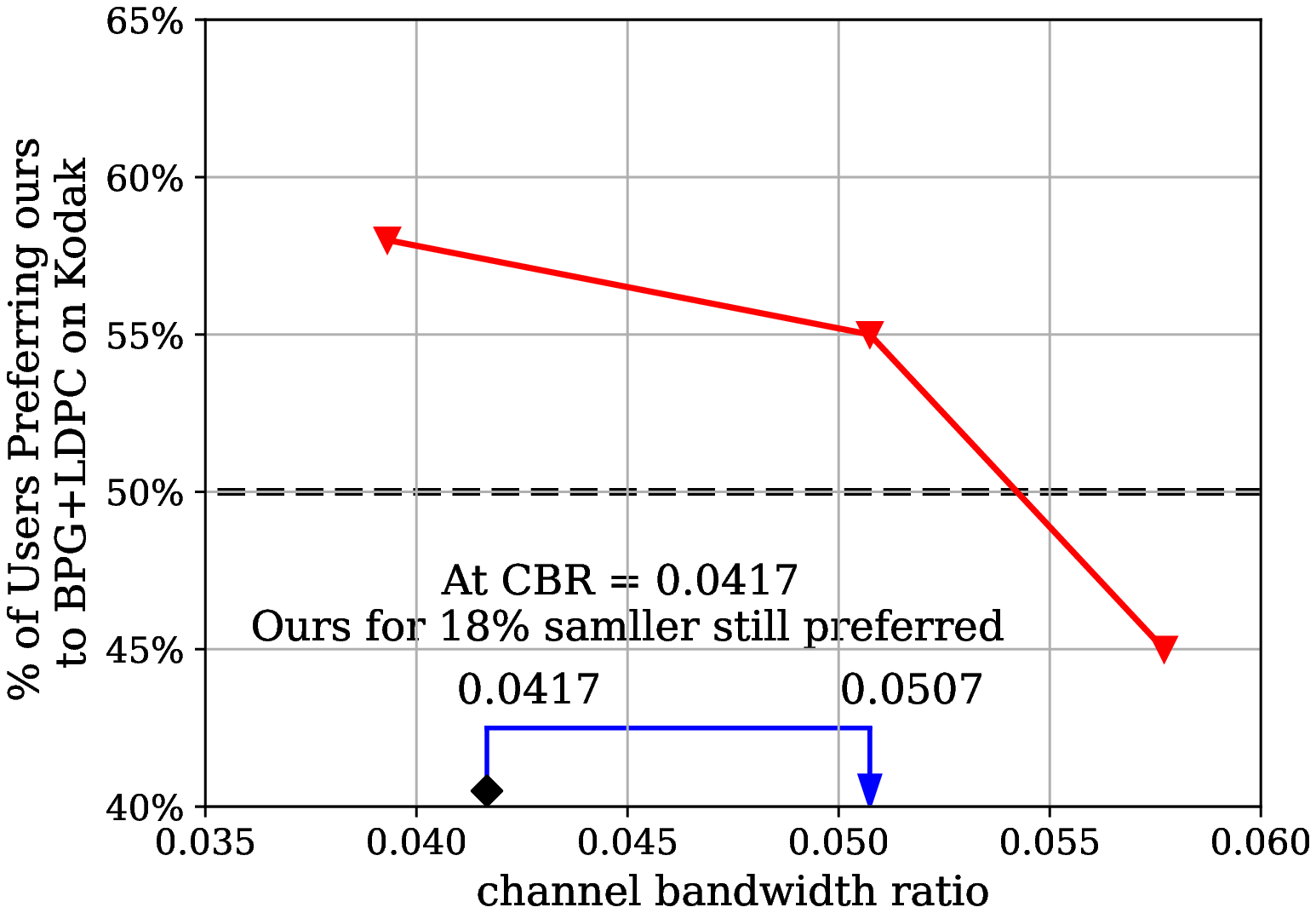}}

	\caption{User study results on Kodak dataset. The red solid line shows the percentage of preferring our scheme to BPG+LDPC at the different average CBR. The blue arrow points from our model to the highest-CBR BPG+LDPC operating point where more than 50\% of users prefer ours, indicating how much more CBR BPG+LDPC uses at that point. }\label{Fig_user_study}
	\vspace{-1em}
\end{figure}

Fig \ref{Fig_user_study} shows the mean preference percentage comparing our proposed models to BPG + LDPC on the Kodak dataset at channel SNR = 10dB. The upper figure illustrates the results at the CBR = 1/48. It can be seen that our method is the preferred scheme even if the CBR cost is 47\% smaller than that of BPG + LDPC. It is well known that the BPG suffers blocky artifacts, which is unfriendly to human visual perception, but our method can well preserve global semantic information and synthesizes realistic texture. The lower figure shows the results at the CBR = 1/24, the overall image quality improvement leads to a relatively smaller gain of our model, but it is still preferred to the BPG + LDPC when saving 18\% channel bandwidth cost. These results verify that our perceptual learned deep JSCC can reconstruct higher visual quality images by using less channel bandwidth cost.  

\begin{figure*}
 \begin{center}
 \hspace{-0.23in}
 \quad
 \subfigure[Original image] {\includegraphics[width=0.24\textwidth]{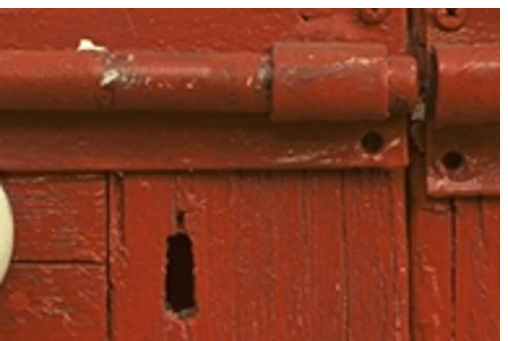}}
 \hspace{-.17in}
 \quad
 \subfigure[Deep JSCC] {\includegraphics[width=0.24\textwidth]{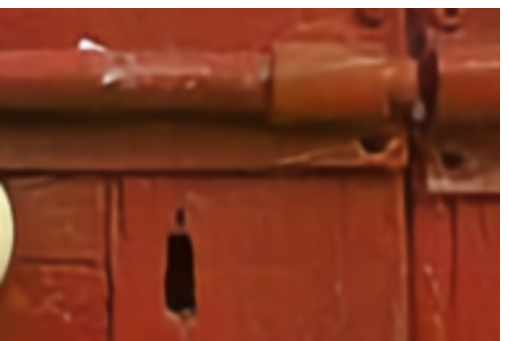}}
 \hspace{-.17in}
 \quad
 \subfigure[$\beta_m = 10^{-3}, \beta_g = 0$]{\includegraphics[width=0.24\textwidth]{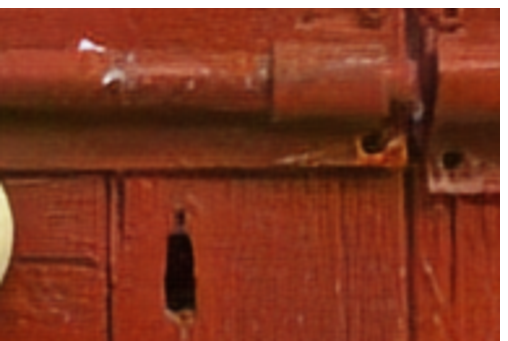}}
 \hspace{-.17in}
 \quad
 \subfigure[$\beta_m = 10^{-4}, \beta_g = 0$] {\includegraphics[width=0.24\textwidth]{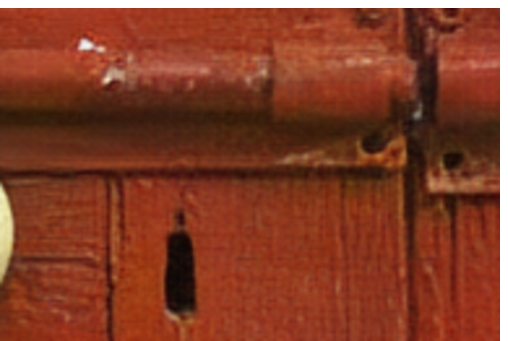}}

 \hspace{-0.0875in}
 \subfigure[$\beta_m = 10^{-5}, \beta_g = 0$] {\includegraphics[width=0.24\textwidth]{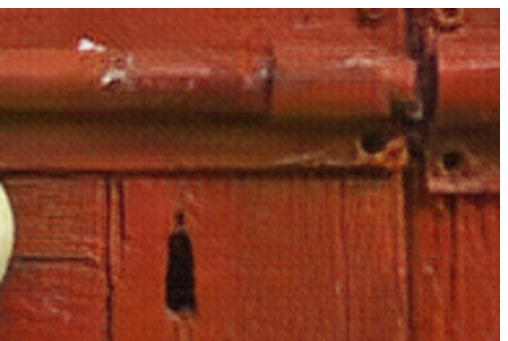}}
 \hspace{-.17in}
 \quad
 \subfigure[$\beta_m = 10^{-5} , \beta_g = 10^{-5}$] {\includegraphics[width=0.24\textwidth]{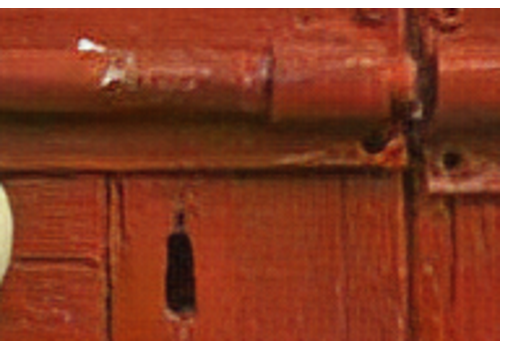}}
 \hspace{-.17in}
 \quad
 \subfigure[$\beta_m = 10^{-5} , \beta_g = 10^{-3}$] {\includegraphics[width=0.24\textwidth]{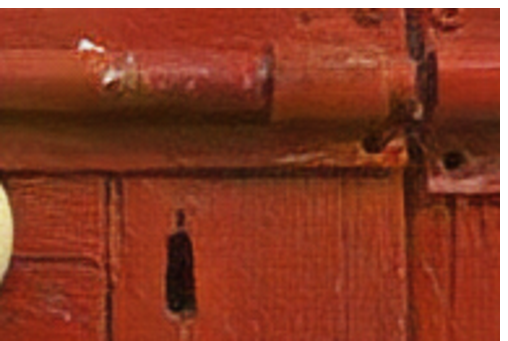}}
 \hspace{-.17in}
 \quad
 \subfigure[$\beta_m = 10^{-5} , \beta_g =10^{-1}$]{\includegraphics[width=0.24\textwidth]{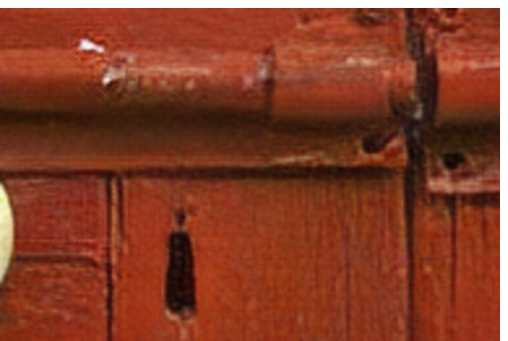}}
\end{center}

\setlength{\abovecaptionskip}{0.cm}
\setlength{\belowcaptionskip}{-0.cm}
\caption{Ablation study on the effect of perceptual loss components over the AWGN channel at SNR = 10dB with the CBR = 1/48.}
\label{Fig_ablation_study}
\vspace{-0.5em}
\end{figure*}

\subsection{Model Variants}

As one type of ablation study, we explore the effects of introducing perceptual loss $d_\text{LPIPS}$ and adversarial loss. We fix the $\beta_p = 1.0$ and vary the $\beta_m$ and $\beta_g$ to observe the visual results. Firstly we set the $\beta_g = 0$ and compare the reconstructed images produced respectively from our models with $\beta_m = 10^{-3}, 10^{-4}, 10^{-5}$ to the baseline deep JSCC model in \cite{DJSCCF}. Fig. \ref{Fig_ablation_study} indicates that when $\beta_m$ decreases, the blurring artifacts are removed gradually. However, it can be discovered that introducing the perceptual loss causes gridding artifacts like Fig. \ref{Fig_ablation_study}(e). This indicates that it still cannot obtain adequate visual pleasing results by only using the perceptual loss $d_{\text{LPIPS}}$ to optimize the deep JSCC model.

We further introduce the adversarial loss respectively setting $\beta_g = 10^{-5}, 10^{-3}, 10^{-1}$ in our model. We can see that the gridding artifacts are removed gradually as $\beta_g$ increases, and many realistic details appear in the image with the help of the discriminator. Our GAN can synthesize some clear local textures to achieve better visual quality. These results verify that our generative deep JSCC can well reconstruct closer to human visual perception images by combing the perceptual loss and adversarial loss.

\section{Conclusion}\label{section_conclusion}

This paper has proposed a new deep JSCC system optimized toward human visual perception for semantic communications. It has introduced perceptual and adversarial losses, enabling the proposed model to capture global semantic information and local texture. Visualization results have demonstrated that our method can produce visually pleasing images instead of blurring/blocky artifacts for low transmission rates. Extensive numerical results and user study have verified that the proposed scheme can generally save communication channel bandwidth costs compared to SOTA engineered image coded transmission systems as well as the standard deep JSCC.

\bibliographystyle{IEEEbib}
\bibliography{Jun_GC_Perceptual}

\end{document}